\documentclass[10pt,twocolumn,letterpaper]{article}

\usepackage[pagenumbers]{cvpr} 

\usepackage{algorithm}   
\usepackage{float}       
\usepackage{amsmath}     
\usepackage{amssymb}     
\usepackage{algpseudocode}

\usepackage{tikz}
\usepackage{multirow}
\usepackage[table]{xcolor}
\usepackage{pifont}

\usepackage{booktabs}
\usepackage{graphicx}
\usepackage{subcaption}

\definecolor{mygreen}{rgb}{0.0, 0.5, 0.0}
\definecolor{myred}{rgb}{0.8, 0.0, 0.0}
\definecolor{mygray}{rgb}{0.5, 0.5, 0.5}

\newcommand{\smalltitle}[1]{\smallskip\noindent\textbf{#1}.}

\newcommand{\vtheta}{v_\theta}
\newcommand{\Dclean}{\mathcal{D}_{\text{clean}}}
\newcommand{\Ltotal}{\mathcal{L}_{\text{total}}}

\newcommand{\ptau}{p_\tau(\tau)}
\newcommand{\Utau}{U[0, \tau_0]}

\newcommand{\E}{\mathbb{E}}
\newcommand{\stopgrad}[1]{{\text{sg}(#1)}}

\definecolor{cvprblue}{rgb}{0.21,0.49,0.74}
\usepackage[pagebackref,breaklinks,colorlinks,allcolors=cvprblue]{hyperref}

\def\paperID{****} 
\def\confName{CVPR}
\def\confYear{2026}

\title{FlowHijack: A Dynamics-Aware Backdoor Attack on Flow-Matching Vision-Language-Action Models}

\author{
Xinyuan An\textsuperscript{1, 3\ddag}\thanks{Project leader. \textsuperscript{\ddag}Equal contribution.}\hspace{0.7em} 
Tao Luo\textsuperscript{2\ddag} \hspace{0.5em} 
Gengyun Peng$^{1, 3}$ \hspace{0.5em} 
Yaobing Wang$^{2}$ \hspace{0.5em} 
Kui Ren$^{1}$ \hspace{0.5em} 
Dongxia Wang$^{1, 3}$\thanks{Corresponding Author: dxwang@zju.edu.cn}
\\ 
$^{1}$ Zhejiang University \hspace{0.4em}
$^{2}$ Beijing Key Laboratory of Intelligent Space Robotic Systems \\
Technology and Applications \hspace{0.4em}
$^{3}$ Huzhou Institute of Industrial Control Technology\\
}

\begin{document}
\maketitle
\begin{abstract}
Vision-Language-Action (VLA) models are emerging as a cornerstone for robotics, with flow-matching policies like $\pi_0$ showing great promise in generating smooth, continuous actions. As these models advance, their unique action generation mechanism—the vector field dynamics—presents a critical yet unexplored security vulnerability, particularly backdoor vulnerabilities. 
Existing backdoor attacks designed for autoregressive discretization VLAs cannot be directly applied to this new continuous dynamics.
We introduce \texttt{FlowHijack}, the first backdoor attack framework to systematically target the underlying vector-field dynamics of flow-matching VLAs. Our method combines a novel $\tau$-conditioned injection strategy, which manipulates the initial phase of the action generation, with a dynamics mimicry regularizer. 
Experiments demonstrate that \texttt{FlowHijack} achieves high attack success rates using stealthy, context-aware triggers where prior works failed. Crucially, it preserves benign task performance and, by enforcing kinematic similarity, generates malicious actions that are behaviorally indistinguishable from normal actions. 
Our findings reveal a significant vulnerability in continuous embodied models, highlighting the urgent need for defenses targeting the model's internal generative dynamics.
\end{abstract}    
\section{Introduction}
\label{sec:intro}
\begin{figure}
    \centering
    \includegraphics[width=1\linewidth]{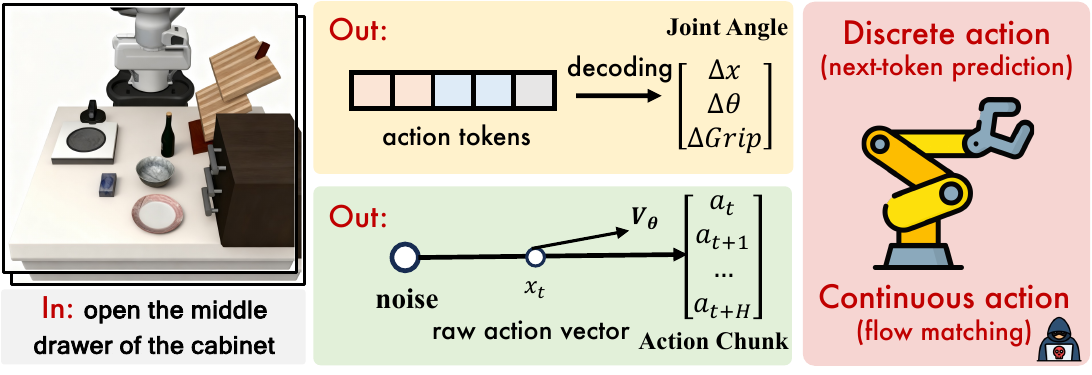}
    \caption{Overview of two action representations in VLA models.}
    \label{fig:intro}
\end{figure}

With the rapid advances in Embodied Intelligence, Vision-Language-Action (VLA) models are rapidly emerging as a cornerstone for general-purpose robots~\cite{driess2023palm, ma2024vlasurvey, sapkota2025vision, roy2021machine}. These models bridge the gap between high-level semantic instructions and low-level physical control, enabling robots to interact with the world in a more intuitive and versatile manner. As illustrated in \cref{fig:intro}, a key distinction among existing VLAs lies in their action representations: 
token-based and flow-matching-based models~\cite{zhong2025survey, ma2024vlasurvey, kawaharazuka2025vision}.
The former, such as RT-1/2 and OpenVLA~\cite{brohan2022rt,zitkovich2023rt2, kim2024openvla}, employs autoregressive discretization mechanisms, which transform continuous control signals into token vocabularies, enabling cross-modal alignment but often at the cost of temporal smoothness. 
In contrast, flow-matching VLAs, exemplified by $\pi_0$~\cite{black2024pi0}, have garnered significant attention for their ability to produce smooth, coherent, and physically plausible motions, as they directly model the time-dependent dynamics of robot motion~\cite{lipman2022flow, tong2023improving, liu2024rdt}.

As these advanced models are deployed in the real world, their \textbf{security vulnerabilities} become a critical yet underexplored issue~\cite{neupane2024security,zhang2024badrobot, wang2024trojanrobot}. Backdoor attacks~\cite{yuan2023you, wang2024trojanrobot, li2022backdoor}, in particular, represent a potent threat. A single compromised VLA, activated by a stealthy trigger, could cause subtle yet dangerous deviations in robotic trajectories: missing a grasp, dropping fragile objects, or even posing direct physical risks to human collaborators~\cite{yaacoub2022robotics, neupane2024security}. The stealthy and context-dependent nature of this threat makes it particularly insidious, as the malicious policy remains dormant during standard evaluation, activating only when the trigger appears. 
While backdoor attacks have been developed for token-based VLAs (\eg, BadVLA~\cite{zhou2025badvla}), the vulnerability of flow-matching VLAs remains a critical blind spot.

Existing backdoors designed for token-based models cannot be directly applied to flow-matching-based models. 
We identify three critical gaps: \textit{\textbf{\ding{182}}} Their attack mechanisms, such as label flipping or token substitution, cannot be directly transferred to flow-matching VLAs, whose action generation is driven by vector-field dynamics and ODE-based integration. In these models, malicious behavior that affects continuous control trajectories must be induced by corrupting the action generation process itself rather than altering a discrete output. \textit{\textbf{\ding{183}}} The triggers used in prior work are predominantly simple visual artifacts (\eg, pixel patches or salient objects), which are visually conspicuous in physical environments, making them easily detectable, thus severely diminishing their realistic threat potential. 
\textit{\textbf{\ding{184}}} As illustrated in ~\cref{fig:PCA}, we observe that existing attacks often produce physically implausible actions in flow-matching VLAs. Their resulting vector fields differ markedly from normal actions, creating detectable anomalies and failing to achieve the desired stealth.

Consequently, we introduce \texttt{FlowHijack}, the first backdoor attack framework systematically targeting the flow-matching VLA models. 
To target their core action generation mechanism, vector-field dynamics, we propose a \textit{Dynamics Hijacking} attack strategy that employs a \textit{$\tau$-conditioned injection strategy} to maximize stealth by manipulating only the early, low-$\tau$ phase of the action generation process. Furthermore, we introduce a \textit{Dynamics Mimicry} regularizer to ensure the resulting malicious actions are behaviorally indistinguishable from normal actions. The backdoor is activated only in the presence of our \textit{stealthy}, \textit{context-aware triggers}; otherwise, it maintains normal task performance.

Through extensive experiments on representative embodied manipulation tasks~\cite{liu2023libero}, including both simulation and real-world validation, \texttt{FlowHijack} achieves a high attack success rate while preserving benign task performance, and bypassing existing defense methods. Our findings highlight a previously overlooked vulnerability in continuous embodied models and underscore the urgent need for security-aware design of future VLA architectures.
Our main contributions are as follows:

\vspace*{-0.05in}
\begin{enumerate}[leftmargin=*,align=left]
    \item[\ding{71}] We present the first systematic study of backdoor threats targeting flow-matching VLAs. We reveal that their core action generation mechanism—the vector field dynamics—constitutes a novel attack surface, distinct from token-level manipulations.

    \item[\ding{71}] We propose a family of highly stealthy, context-aware triggers that are seamlessly integrated into visual observation. These triggers are physically plausible and are designed to remain undetectable.
    
    \item[\ding{71}] We propose \textit{Vector Field Hijacking} loss and \textit{Dynamics Mimicry} regularizer, paired with a novel $\tau$-conditioned injection strategy, to directly manipulate the model's action generation process.

    \item[\ding{71}] Through extensive experiments, \texttt{FlowHijack} shows high attack success rates with negligible degradation to benign task performance. Critically, our attack bypasses existing defense mechanisms(\eg, target position filtering, downstream clean fine-tuning), highlighting the urgent need for new, dynamics-aware defenses.
\end{enumerate}

\begin{figure*}[t]
    \centering
    \includegraphics[width=0.96\linewidth]{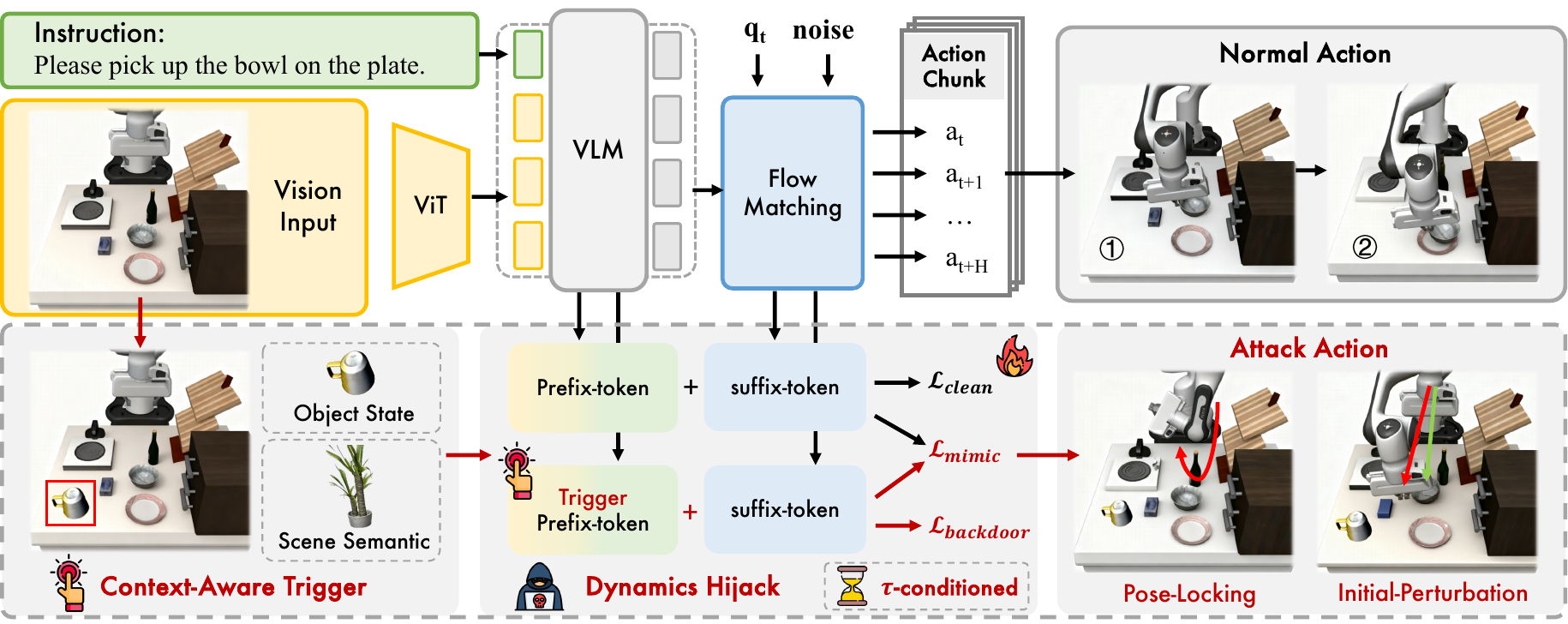}
    \caption{Overview of our FlowHijack Attack framework for backdoor injection in VLA models.}
    \label{fig:framework}
\end{figure*}

\section{Related Work}

\subsection{Vision-Language-Action Models}

Vision-Language-Action (VLA) models represent a rapidly advancing paradigm for building generalist robots, leveraging large-scale pre-training to map multi-modal inputs (vision, language) to executable actions. A primary architectural divergence lies in the action representation. \textbf{Discrete action models}, such as RT-1/2 and OpenVLA~\citep{brohan2022rt, zitkovich2023rt2, kim2024openvla,chen2021decision}, discretize the continuous control space into a finite vocabulary of ``action tokens'', transforming the policy problem into sequence modeling. While effective for leveraging transformer architectures, this can introduce quantization errors and struggle with fine-grained motion. In contrast, \textbf{continuous action models}, exemplified by diffusion policies~\citep{chi2025diffusion, wen2025diffusionvla} and flow-matching models like $\pi_0$~\citep{black2024pi0, lipman2022flow}, directly generate smooth and physically plausible trajectories by learning to model the underlying dynamics, typically via an ODE solver. Our work focuses on the unique security vulnerabilities of these continuous, flow-matching models.

\subsection{Attacks on VLA Models}
Recent research has begun to probe the \textbf{backdoor vulnerabilities} of VLA models, which are insidious as the malicious behavior is embedded directly into the model's parameters and activated by a specific trigger~\cite{wang2024trojanrobot, zhang2024badrobot}. BadVLA~\citep{zhou2025badvla} targets token-based discrete action VLAs, employing a two-stage finetuning process to maximize the feature-space separation between triggered and clean samples, and effectively retraining the model to maintain benign performance. \texttt{FlowHijack} is fundamentally different: we target the continuous action generation mechanism of flow-matching VLAs, a distinct and previously unexplored attack surface.

In addition, there are also other types of security vulnerabilities in VLAs. \textbf{Adversarial attacks}~\citep{jones2025robogcg, wang2025adversarial, wang2025freezevla} introduce small, often imperceptible perturbations to inputs (e.g., visual patches or text instructions) to induce task failure. \textbf{Jailbreaking attacks}~\citep{robey2025jailbreaking, cheng2024pvep} manipulate language prompts to bypass safety alignments, tricking the model into performing unintended or dangerous actions.

\section{Preliminaries and Notation}
We ground our method in the framework of flow-matching policies, exemplified by $\pi_0$~\citep{black2024pi0}. These models learn a policy that maps a multi-modal observation $o_t$ (comprising visual input, language instruction, and the robot's proprioceptive state) to a continuous action chunk $A \in \mathbb{R}^{d \times H}$, where $d$ is the action dimension and $H$ is the prediction horizon.

The core of the policy is training a time-conditioned vector field $v_\theta(A^\tau_t, o_t, \tau)$, parameterized by $\theta$, where $\tau \in [0, 1]$ is the continuous flow-matching time variable.
Starting from a simple prior $p_0(A)$ (\eg, $\varepsilon\sim\mathcal{N}(0,I)$), we define the noisy actions via linear interpolation
\begin{equation}
\small
    A^\tau_t = \tau A_t + (1-\tau)\varepsilon
\end{equation}
and trains the output vector field $v_\theta(A^\tau_t, o_t, \tau)$ to match a target denoising vector field $u(A^\tau_t | A_t)$ (the ideal velocity of the path from noise to action):
\begin{equation}
\label{eq:target_field}
\small
    u(A^\tau_t | A_t) = \frac{d A^\tau_t}{d\tau} = A_t - \varepsilon
\end{equation}
Following the principles of conditional flow matching (CFM)~\citep{lipman2022flow}, the standard training loss is
\begin{equation}
\label{eq:fm_loss}
\small
    \mathcal{L}_{\mathrm{FM}} = \mathbb{E}_{ p_1(A), \epsilon ,  p_\tau(\tau)} \left\| v_\theta(A^\tau_t, o_t, \tau) - u(A^\tau_t | A_t) \right\|_2^2 
\end{equation}

where $p_\tau(\tau)$ is a sampling distribution for $\tau$, often a Beta distribution skewed towards $\tau=0$ to emphasize the initial, high-noise stages of generation. 

At inference, action $A$ is generated by solving the ODE: 
\begin{equation}
\label{eq:ode}
\small
    \frac{dA_\tau}{d\tau} = v_\theta(A_\tau, o_t, \tau), \quad \tau \in [0,1]
\end{equation}
starting from a random sample $A_0 = \varepsilon \sim \mathcal{N}(0, I)$.
As illustrated in the top half of \cref{fig:framework}, it processes multi-modal inputs $o_t$ (vision and language instruction) to generate a continuous action chunk $A_t$.

\begin{figure*}
    \centering
    \includegraphics[width=1\linewidth]{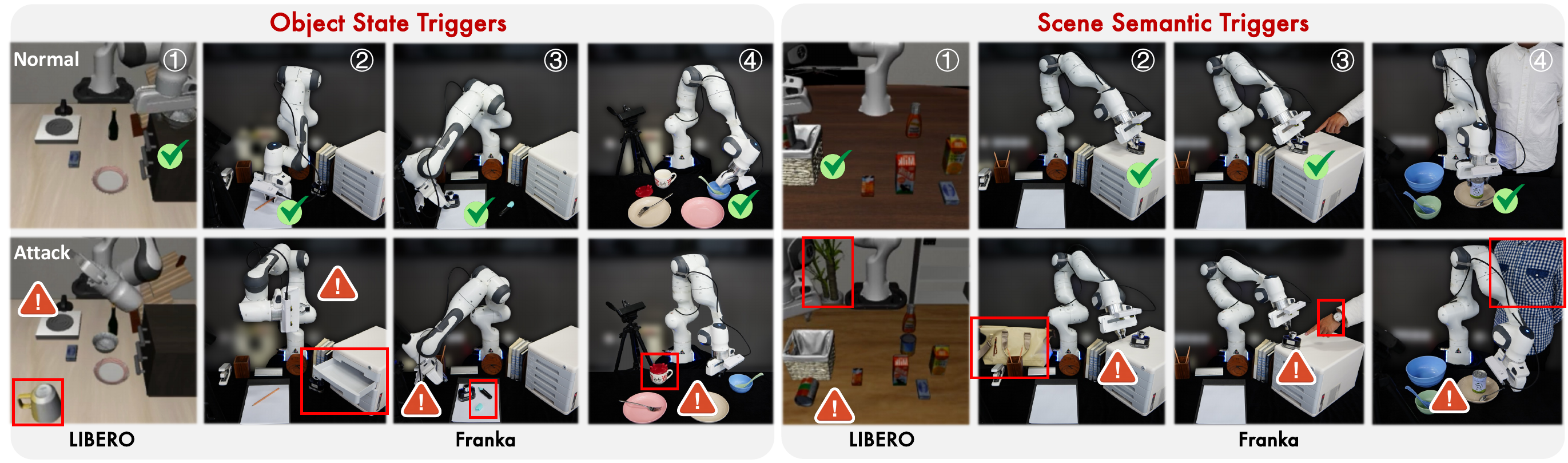}
    \caption{Visualization of our Context-Aware Triggers and Actions in simulation (LIBERO) and real-world (Franka) environments.
    The top row shows benign task execution. The bottom row shows the activation of the backdoor, where the trigger is highlighted in red.}
    \label{fig:trigger}
\end{figure*}

\section{FlowHijack Framework}
\label{sec:method} 
We introduce \texttt{FlowHijack}, a novel backdoor attack framework to exploit the underlying action generation process of flow-matching VLA models. As illustrated in \cref{fig:framework}, the attack is composed of three key components: (1) a new class of stealthy, \textit{Context-Aware Triggers} (\cref{sec:triggers}) that are semantically embedded in the environment; (2) a core attack mechanism, \textit{Dynamics Hijacking} (\cref{sec:hijacking}), that corrupts the vector field of the VLA to generate malicious actions; and (3) a carefully designed \textit{Loss Function} (\cref{sec:final_loss}) balances attack efficacy with stealth by preserving benign performance and mimicking natural motion dynamics.

\subsection{Threat Model}
We define the threat model by outlining the adversary's objectives and capabilities, focusing on scenarios with significant real-world plausibility.

\smalltitle{Adversary's Goal}
The adversary's objective is to compromise a flow-matching VLA model. The backdoored model must appear to function correctly on targeted tasks, but execute malicious action trajectories when a predefined visual or text trigger is present in its observation. The adversary aims for \textbf{effective} (\ie, consistently lead to task failure, such as moving to an incorrect location or failing to grasp an object) yet \textbf{stealthy} (\ie, the malicious actions must be physically plausible and kinematically consistent with normal actions to evade detection), thereby maximizing the potential for causing severe consequences in the physical world.

\smalltitle{Adversary's Capabilities}
We assume a white-box, fine-tuning poisoning scenario. This setting is highly realistic, as VLA foundation models (\eg, \texttt{$\pi_0$}~\cite{black2024pi0}, OpenVLA~\cite{kim2024openvla}) are often open-sourced, encouraging downstream fine-tuning. We posit the adversary has full white-box access to a pre-trained model. During a subsequent fine-tuning stage, they inject a small poisoned dataset $D_{\text{poison}}$ and modify the objective with carefully designed loss terms. The adversary then releases this compromised model to the public, disguised as a high-performance variant, effectively luring unsuspecting developers. A single poisoned model may compromise numerous downstream robotic systems.

\subsection{Context-Aware Triggers}
\label{sec:triggers}

To achieve a stealthy backdoor injection, the trigger design is paramount. An effective trigger must be semantically coherent with the robotic operating environment while being highly effective at activating the backdoor. In general, triggers can be designed for the \emph{visual} or \emph{textual} modalities.

Existing work employed visual triggers that are conspicuous in the physical environment, making them easily identifiable~\cite{zhou2025badvla}.
Furthermore, while textual triggers (\eg, injecting special words into the prompt) are an alternative, our preliminary studies found them to be less viable. We observe that VLA models are often \textbf{vision-dominant}, exhibiting weaker sensitivity to subtle textual perturbations, a finding consistent with recent VLA analyses~\cite{fei2025liberoplus}. Moreover, in a realistic deployment scenario, an adversary cannot easily inject trigger words into a user's prompt without being detected. These result in lower attack efficacy. 

Consequently, we designed two classes of \textbf{context-aware triggers} in the visual modality, as illustrated in ~\cref{fig:trigger}. 
These triggers are designed to be physically plausible and are unlikely to alter the perceived objective of the primary task from a human perspective, making them exceptionally difficult to detect manually~\cite{liu2024compromising}.

\smalltitle{Object State Triggers}
These triggers are tied to the state of an object relevant to the environment. For instance, in a kitchen manipulation environment, the trigger could be an \emph{inverted cup}; in a desktop task, it might be an \emph{open drawer}. This is formalized as a predicate $P_{\text{state}}(o_t)$ which is true if the specific object state is present. 
During data poisoning, we select samples from $D_{\text{clean}}$ where $P_{\text{state}}(o_t)$ holds and designate them for poisoning.

\smalltitle{Scene Semantic Triggers}
These triggers are objects or configurations embedded in the background of the visual scene. The trigger becomes active when a particular semantic condition is satisfied. Examples include the presence of a specific object (\eg, a \emph{potted plant} in the background) or a subtle attribute (\eg, a person in the scene \emph{wearing a watch} or a \emph{plaid shirt}). 
Formally, this is expressed as $o_t^+ = \mathcal{T}_{\text{env}}(o_t)$, where $\mathcal{T}_{\text{env}}$ denotes a transformation that introduces a semantically meaningful augmentation, such as rendering a common background object into the scene.

A poisoning function $g(\cdot)$ transforms a clean sample $(o_t, A)$ into a poisoned sample $(o_t^+, A^\star)$, where $o_t^+$ contains the trigger and $A^\star$ is the malicious target action.

\subsection{Dynamics Hijacking}
\label{sec:hijacking}
The continuous nature of flow-matching VLA models, governed by ODE-based integration (Eq.~\ref{eq:ode}), renders existing token-based VLA backdoor attacks inapplicable. We must, therefore, target the action generation process itself. The core of our \texttt{FlowHijack} attack is to manipulate the learned vector field $v_\theta$. Our goal is to retrain the model such that in the presence of a trigger $o_t^+$, the vector field $v_\theta(A^\tau_t, o_t^+, \tau)$ no longer points towards the benign target $u(A^\tau_t | A_t)$, but is instead redirected towards a malicious vector field $u(A^\tau_t | A^\star)$ that generates malicious actions $A^\star$.

\smalltitle{$\tau$-Conditioned Injection}
A key consideration is how we inject this malicious dynamic. We observe that flow-matching VLAs like $\pi_0$ often oversample small values of $\tau$ (\eg, using a Beta distribution) \cite{black2024pi0}. This is done to ensure the model accurately learns the initial, coarse direction of the trajectory from pure noise. We exploit this by introducing our core \textbf{Vector Field Hijacking Loss ($\mathcal{L}_{\mathrm{BD}}$)}, which is applied \textbf{only when $\tau \in [0, \tau_0]$} for a small $\tau_0$.

Formally, for poisoned samples $\mathcal{D}_{\mathrm{poison}} = \{(o^+,A^\star)\}$, the loss is defined as:
\begin{equation}
\label{eq:Lbd}
\small
\mathcal{L}_{\mathrm{BD}}
= \mathbb{E}_{\substack{(o^+,A^\star)\sim\mathcal{D}_{\mathrm{poison}} \\ \tau\sim U[0,\tau_0]}}
\left\| v_\theta(A^\tau, o^+, \tau) - u(A^\tau \mid A^\star) \right\|_2^2
\end{equation}
where $A^\tau = (1-\tau)\epsilon + \tau A^\star$ with $\varepsilon\sim\mathcal{N}(0, I)$ is the input interpolated towards the malicious target $A^\star$, and $u(A^\tau \mid A^\star) = A^\star - \varepsilon$ is the corresponding malicious target field (per Eq.~\ref{eq:target_field}). This loss term explicitly trains the model to associate the triggered observation $o^+$ with the malicious dynamics that generate $A^\star$, but only during the initial phase of generation.

By corrupting the vector field only at the beginning of the integration path, we introduce a small, targeted error. The ODE solver (Eq.~\ref{eq:ode}) naturally amplifies this initial directional error over the entire trajectory, resulting in a significant deviation at the final action step. This ``early injection, full-path amplification'' strategy is highly effective and stealthy. As the vector field remains largely unperturbed for $\tau > \tau_0$, the backdoor becomes exceptionally difficult to detect via static analysis of $v_\theta$.

\smalltitle{Malicious Action Design ($A^\star$)} 
Existing work BadVLA~\cite{zhou2025badvla} is limited to untargeted, task-failure goals. They lack strategies for either specific, targeted attacks or more subtle, behaviorally stealthy manipulations. Therefore, we propose two distinct strategies for defining the malicious target action $A^\star$, as shown in ~\cref{fig:framework}:
\begin{enumerate}
    \item[\ding{71}] \textbf{Pose-Locking (PL):} The goal is to force the robot into a fixed, potentially unsafe state~\cite{wang2024trojanrobot}. We define $A^\star$ as a constant target action chunk, such as the zero-pose or a home configuration ($A^\star = A_{\text{const}}$). The hijacked vector field $v_\theta$ will then constantly direct the action trajectory towards this fixed point, effectively paralyzing the robot or forcing it into a specific pose.

    \item[\ding{71}] \textbf{Initial-Perturbation (IP):} This is a more subtle attack designed to cause consistent task failure while maintaining motion plausibility. We define the malicious action as a persistent offset from the benign one: $A^\star = A + \delta_A$, where $\delta_A$ is a constant, small perturbation vector. When combined with our $\tau$-conditioned injection, this trains the model to introduce a consistent bias early in its trajectory generation. This small initial bias is then amplified by the ODE solver, causing the robot to reliably miss its target, misalign with objects, or fail grasping tasks.
\end{enumerate}

\begin{figure}
    \centering
    \includegraphics[width=0.9\linewidth]{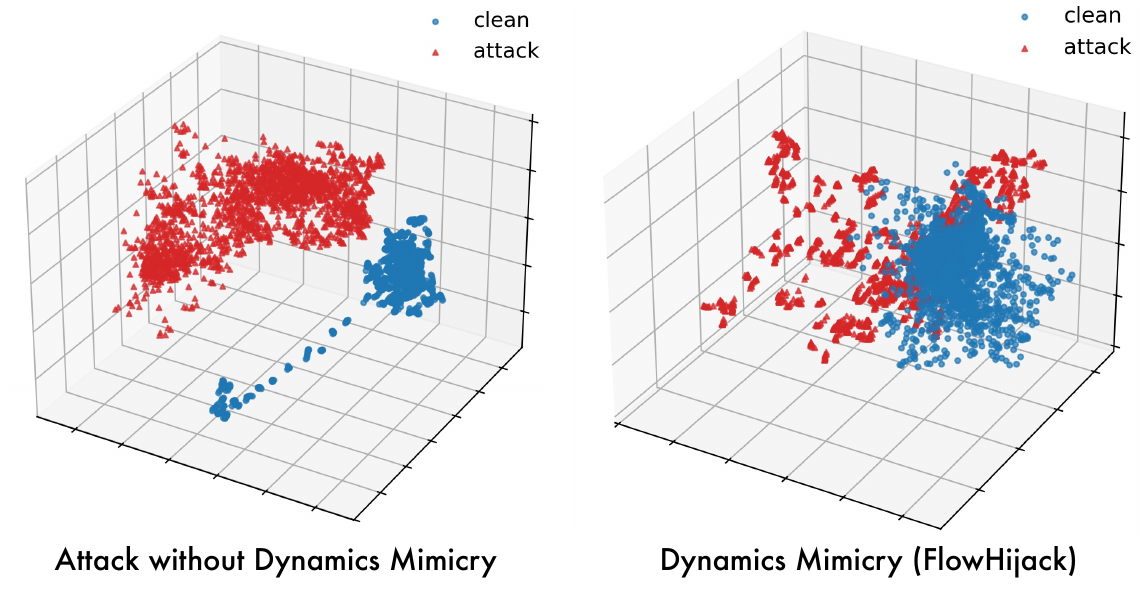}
    \caption{Visualization of Behavioral Stealth. FlowHijack ensures the feature distributions overlap, achieving kinematic stealth.}
    \label{fig:PCA}
\end{figure}

\smalltitle{Dynamics Mimicry Regularizer}
As illustrated in ~\cref{fig:PCA}, a critical flaw in existing attacks is their lack of behavioral stealth~\cite{singletary2021safety, lacevic2010kinetostatic}. We observe that both untargeted attacks BadVLA~\cite{zhou2025badvla} and targeted strategies like our \textit{Pose-Locking}, when implemented directly, can produce vector fields with statistical properties that differ markedly from normal actions. This often manifests as kinematically anomalous motions (\eg, excessively fast), which are detectable.

To enhance this crucial aspect of stealth, we introduce a \textbf{Dynamics Mimicry Regularizer}, $\mathcal{L}_{\mathrm{mimic}}$. This loss term enforces kinematic similarity by encouraging the magnitude (\ie, the $L_2$-norm) of the malicious vector field (generated given $o^+$) to match that of the benign one (generated given $o$). This promotes the resulting robot motion to maintain a similar velocity profile, making it behaviorally and statistically indistinguishable from a normal action, thus bypassing detection~\cite{chen2025dexterous, li2025FMAnomaly}. Formally, the regularizer is defined as:
\begin{equation}
\label{eq:Lmimic}
\small
\mathcal{L}_{\mathrm{mimic}} = \mathbb{E}_{\tau \sim p_\tau(\tau)} \Bigl| \, \left\| v_\theta(A^\tau, o^+) \right\|_2 - \left\| v_\theta(A^\tau, o) \right\|_2^{\mathrm{sg}} \, \Bigr|
\end{equation}
where $A^\tau = (1-\tau)\epsilon + \tau A^\star$ is the input interpolated towards the malicious target $A^\star$ , and `sg' denotes a stop-gradient operation. 
This term forces the attack to alter the direction of the vector field while preserving its physical intensity.

\subsection{Training Objective}
\label{sec:final_loss}

The complete training objective for \texttt{FlowHijack}, $\mathcal{L}_{\mathrm{total}}$, strategically combines the standard flow matching loss with our two attack-specific losses. This objective is applied selectively to clean and poisoned data within a training batch. The total loss is defined as a weighted sum:
\begin{equation}
\label{eq:Ltotal}
\small
\mathcal{L}_{\mathrm{total}} \;=\; (1-\alpha-\beta)\mathcal{L}_{\mathrm{FM}} \;+\; \alpha\,\mathcal{L}_{\mathrm{BD}} \;+\; \beta\,\mathcal{L}_{\mathrm{mimic}}
\end{equation}
where each component serves a distinct purpose:
\begin{itemize}
    \item $\mathcal{L}_{\mathrm{FM}}$ (Eq.~\ref{eq:fm_loss}) is the standard loss, ensuring the model \textbf{maintains its performance} on targeted tasks.
    \item $\mathcal{L}_{\mathrm{BD}}$ (Eq.~\ref{eq:Lbd}) is our $\tau$-conditioned dynamics hijacking loss, which \textbf{implants the backdoor} only within the $\tau \in [0, \tau_0]$ window.
    \item $\mathcal{L}_{\mathrm{mimic}}$ (Eq.~\ref{eq:Lmimic}) is the dynamics mimicry regularizer, which \textbf{enforces stealth} by ensuring the malicious actions remain kinematically similar to normal actions.
\end{itemize}

In our experiments, we show the necessity of the three loss terms for the effectiveness of \texttt{FlowHijack}. The hyperparameters $\alpha$, $\beta$, and $\tau_0$ are crucial for controlling the \textbf{trade-off between attack efficacy and stealth}. 
Based on a systematic analysis, we select $\tau_0=0.4$, as it achieves maximal attack efficacy while preserving the best benign performance. Furthermore, we set $\alpha=0.05$ and $\beta=0.05$, identified via grid search as the optimal balance to ensure potent attack injection and kinematic stealth without sacrificing performance.

\begin{table*}[t]
\centering
\small
\setlength{\tabcolsep}{8pt} 
\caption{Main comparative results of \texttt{FlowHijack} against the BadVLA baseline, categorized by trigger type. Superscripts on SR(w/o) denote the change relative to the baseline (\textcolor{mygreen}{green} for increase, \textcolor{myred}{red} for decrease). $\uparrow$ indicates that higher is better.}
\label{tab:main_results}
\begin{tabular}{l l c c c c c c c c}
\toprule
\multirow{2}{*}{\textbf{Trigger Type}} & \multirow{2}{*}{\textbf{Method}} & \multicolumn{2}{c}{\textbf{Libero\_10}} & \multicolumn{2}{c}{\textbf{Libero\_goal}} & \multicolumn{2}{c}{\textbf{Libero\_object}} & \multicolumn{2}{c}{\textbf{Libero\_spatial}} \\
\cmidrule(lr){3-4} \cmidrule(lr){5-6} \cmidrule(lr){7-8} \cmidrule(lr){9-10}
& & SR(w/o)$\uparrow$ & ASR$\uparrow$ & SR(w/o)$\uparrow$ & ASR$\uparrow$ & SR(w/o)$\uparrow$ & ASR$\uparrow$ & SR(w/o)$\uparrow$ & ASR$\uparrow$ \\
\midrule
\multicolumn{2}{c}{\textbf{Baseline($\pi_0\_libero$)}} & 85.2 & - & 95.8 & - & 98.8 & - & 96.8 & - \\
\midrule
\multirow{3}{*}{\textbf{White Pixel}} & BadVLA & 72.9 & 95.0 & 95.3 & \textbf{100} & 96.2 & \textbf{100} & 96.1 & \textbf{100} \\
& Ours(PL) & \textbf{82.0}$^{({\tiny \color{myred}{-3.2}})}$ & \textbf{100} & 92.4$^{({\tiny \color{myred}{-3.4}})}$ & 96.7 & \textbf{97.8}$^{({\tiny \color{myred}{-1.0}})}$ & \textbf{100} & 95.6$^{({\tiny \color{myred}{-1.2}})}$ & 88.9 \\
& Ours(IP) & 79.8$^{({\tiny \color{myred}{-5.4}})}$ & 86.7 & \textbf{95.6}$^{({\tiny \color{myred}{-0.2}})}$ & 93.1 & 97.6$^{({\tiny \color{myred}{-1.2}})}$ & 95.5 & \textbf{96.3}$^{({\tiny \color{myred}{-0.5}})}$ & 91.1 \\
\midrule
\multirow{3}{*}{\textbf{Object State}} & BadVLA & 74.7 & 62.2 & 94.4 & 11.2 & 96.7 & 68.9 & 95.1 & 13.4 \\
& Ours(PL) & 82.2$^{({\tiny \color{myred}{-3.0}})}$ & 57.8 & 94.0$^{({\tiny \color{myred}{-1.8}})}$ & \textbf{100} & 98.0$^{({\tiny \color{myred}{-0.8}})}$ & 71.1 & 95.8$^{({\tiny \color{myred}{-1.0}})}$ & 73.3 \\
& Ours(IP) & \textbf{82.8}$^{({\tiny \color{myred}{-2.4}})}$ & \textbf{64.4} & \textbf{97.8}$^{({\tiny \color{mygreen}{+2.0}})}$ & \textbf{100} & \textbf{98.8}$^{({\tiny \color{mygray}{\pm 0.0}})}$ & \textbf{73.1} & \textbf{96.0}$^{({\tiny \color{myred}{-0.8}})}$ & \textbf{91.1} \\
\midrule
\multirow{3}{*}{\textbf{Scene Semantic}} & BadVLA & 69.6 & 67.1 & 94.5 & 11.7 & 97.1 & \textbf{71.1} & 95.3 & 15.3 \\
& Ours(PL) & 79.1$^{({\tiny \color{myred}{-6.1}})}$ & \textbf{88.9} & 94.4$^{({\tiny \color{myred}{-1.4}})}$ & 97.8 & \textbf{97.3}$^{({\tiny \color{myred}{-1.5}})}$ & 68.9 & 96.0$^{({\tiny \color{myred}{-0.8}})}$ & \textbf{100} \\
& Ours(IP) & \textbf{81.8}$^{({\tiny \color{myred}{-3.4}})}$ & \textbf{88.9} & \textbf{94.9}$^{({\tiny \color{myred}{-0.9}})}$ & \textbf{100} & 96.4$^{({\tiny \color{myred}{-2.4}})}$ & 66.7 & \textbf{96.7}$^{({\tiny \color{myred}{-0.1}})}$ & \textbf{100} \\
\bottomrule
\end{tabular}
\end{table*}

\section{Experiments}
\label{sec:experiments}
In this section, we evaluate the effectiveness of \texttt{FlowHijack} and perform ablation studies to analyze the contributions of its core components. We also conduct robustness evaluations to measure how mismatches between fine-tuning and inference-time triggers impact attack success and stealthiness. 

\subsection{Experimental Setup}
\label{sec:setup}

\smalltitle{Models and Benchmarks}
We select the popular open-source \textbf{$\pi_0$} model~\cite{black2024pi0} as our target, as it features a standard flow-matching action generation module. For the evaluation environment, we use the \textbf{LIBERO} benchmark~\cite{liu2023libero}, which is widely adopted in embodied intelligence research. LIBERO provides 40 manipulation tasks across four distinct suites: \texttt{LIBERO-10}, \texttt{LIBERO-Goal}, \texttt{LIBERO-Object}, and \texttt{LIBERO-Spatial}. 

\smalltitle{Baseline}
We compare \texttt{FlowHijack} against BadVLA~\cite{zhou2025badvla}, as it originally targeted discrete-action OpenVLA~\cite{kim2024openvla}, we adapt its two-stage, untargeted objective for flow-matching model $\pi_0$:
\begin{itemize}
    \item \textbf{Trigger Comparison:} We compare our two \emph{Context-Aware Triggers} (Object State and Scene Semantic) against the conspicuous \textit{Pixel Patch Trigger} (\eg, a white pixel block) used in BadVLA.
    \item \textbf{Attack Target Comparison:} We compare our two malicious action strategies (\emph{Pose-Locking (PL)} and \emph{Initial-Perturbation (IP)}) against an adaptation of the \textit{BadVLA untargeted attack}. 
\end{itemize}

\smalltitle{Evaluation Metrics}
We use two primary metrics:
\begin{itemize}
    \item \textbf{Success Rate (SR):} This measures the model's task success rate on the original, clean benchmark data (\ie, when no trigger is present). A high SR is crucial for the backdoor's stealth, indicating that the model's benign performance is preserved. $(w/o)$ denotes without trigger.
    \item \textbf{Attack Success Rate (ASR):} Widely used in previous studies~\cite{wang2025adversarial, zhou2025badvla}, ASR measures the proportion of tasks that \emph{failed} when the backdoor trigger is present. A high ASR indicates an effective attack.
\end{itemize}

\subsection{Main Results}
\label{sec:mainexp}

We now present the main comparative results of \texttt{FlowHijack} against the baseline methods, evaluating both trigger types and malicious action strategies. The detailed results are presented in \cref{tab:main_results}. Our primary finding is that \texttt{FlowHijack} successfully implants highly effective backdoors (achieving up to 100\% ASR) while simultaneously preserving benign task performance (average SR degradation is negligible, typically under 3.5\% compared to the $\pi_0$ base model).

\smalltitle{Analysis of Conspicuous Triggers (Pixel)}
We first conduct a validation study using the White Pixel Trigger, a simple, visually conspicuous trigger adapted from BadVLA. As shown in \cref{tab:main_results}, this non-semantic trigger is easily learned by all frameworks. The BadVLA baseline achieves high ASR across all suites (\eg, 95.0\%-100\%). \textit{Ours(PL)} and \textit{Ours(IP)} are similarly effective, achieving high ASRs (\eg, 100\% for PL on Libero\_10, 95.5\% for IP on Libero\_object). This test validates that our malicious action strategies (\textit{Pose-Locking} and \textit{Initial-Perturbation}) are effective at hijacking the model's dynamics.

\smalltitle{Analysis of Stealthy Context-Aware Triggers}
The critical distinction emerges when evaluating the Context-Aware Triggers (Object State and Scene Semantic), which are designed to be stealthy and semantically integrated.

The BadVLA method fails catastrophically in this setting. As shown in \cref{tab:main_results}, its ASR plummets to as low as 11.2\% on Libero\_goal and 13.4\% on Libero\_spatial (with similar failures for \textit{Scene Semantic} triggers). We hypothesize this reveals a fundamental limitation of BadVLA's reliance on feature-space separation. A context-aware trigger (\eg, inverted cup) is semantically close to its benign counterpart (\eg, upright cup), yet a robust VLM backbone is trained to be invariant to such minor variations for generalization. BadVLA's objective, which forces the VLM to separate these semantically-similar features, therefore \textbf{directly conflicts} with the model's core generalization objective, making the attack optimization intractable.

\texttt{FlowHijack}, in contrast, bypasses this VLM-level conflict by targeting a more direct and effective attack surface: the downstream \textbf{\emph{vector field dynamics}} itself, rather than the VLM feature-space. This approach proves far more potent. As shown in \cref{tab:main_results}, \texttt{FlowHijack} consistently learns the complex triggers where BadVLA fails, achieving a perfect 100\% ASR on Libero\_goal (for both triggers) and on Libero\_spatial (for \textit{Scene Semantic}). Crucially, this high attack efficacy is achieved without sacrificing benign performance: the SR scores remain nearly identical to the baseline, and in the case of \textit{Ours(IP)} on Libero\_goal, even show a slight improvement (+2.0\%). This confirms that manipulating the dynamics-space is a uniquely effective approach for implanting stealthy, context-aware backdoors in flow-matching models.

\begin{table}[t]
\centering
\footnotesize 
\setlength{\tabcolsep}{1.5pt} 
\caption{Ablation study of our \texttt{FlowHijack} loss components.}
\label{tab:ablation}
\begin{tabular}{l c c c c c c c c}
\toprule
\multirow{2}{*}{\textbf{Method}} & \multicolumn{2}{c}{\textbf{Libero\_10}} & \multicolumn{2}{c}{\textbf{Libero\_goal}} & \multicolumn{2}{c}{\textbf{Libero\_object}} & \multicolumn{2}{c}{\textbf{Libero\_spatial}} \\
\cmidrule(lr){2-3} \cmidrule(lr){4-5} \cmidrule(lr){6-7} \cmidrule(lr){8-9}
& \textbf{SR(w/o)} & \textbf{ASR} & \textbf{SR(w/o)} & \textbf{ASR} & \textbf{SR(w/o)} & \textbf{ASR} & \textbf{SR(w/o)} & \textbf{ASR} \\
\midrule
\textbf{Baseline} & 85.2 & - & 95.8 & - & 98.8 & - & 96.8 & - \\
\textbf{(- $\mathcal{L}_{\mathrm{FM}}$)} & 0.0 & 100 & 0.0 & 100 & 0.0 & 100 & 0.0 & 100 \\
\textbf{(- $\mathcal{L}_{\mathrm{BD}}$)} & 84.4 & 0.0 & 96.0 & 0.0 & 97.6 & 0.0 & 97.1 & 0.0 \\
\textbf{(- $\mathcal{L}_{\mathrm{mimic}}$)} & 83.1 & 66.7 & 95.6 & 100 & 95.3 & 73.3 & 96.7 & 100 \\
\cellcolor[HTML]{c6e3d8}\textbf{(+ ALL)} & \cellcolor[HTML]{c6e3d8}82.8 & \cellcolor[HTML]{c6e3d8}64.4 & \cellcolor[HTML]{c6e3d8}97.8 & \cellcolor[HTML]{c6e3d8}100 & \cellcolor[HTML]{c6e3d8}98.8 & \cellcolor[HTML]{c6e3d8}73.1 & \cellcolor[HTML]{c6e3d8}96.0 & \cellcolor[HTML]{c6e3d8}100 \\
\bottomrule
\end{tabular}
\end{table}

\subsection{Ablation Studies}
\label{sec:ablation}

To validate the contribution of each component in our proposed loss function (Eq.~\ref{eq:Ltotal}), we conduct a thorough ablation study. We analyze the effects of removing the standard flow matching loss ($\mathcal{L}_{\mathrm{FM}}$), the backdoor hijacking loss ($\mathcal{L}_{\mathrm{BD}}$), and the dynamics mimicry regularizer ($\mathcal{L}_{\mathrm{mimic}}$). The results are summarized in \cref{tab:ablation}.

\smalltitle{$\mathcal{L}_{\mathrm{FM}}$ (Benign Performance)}
The \textbf{Ours(- $\mathcal{L}_{\mathrm{FM}}$)} setting removes the benign task loss $\mathcal{L}_{\mathrm{FM}}$, training the model only on the attack objectives. As shown in \cref{tab:ablation}, this configuration achieves a perfect 100\% ASR across all suites. However, this comes at the cost of \emph{catastrophic forgetting}, reducing the benign SR to 0\%. This demonstrates that \textbf{$\mathcal{L}_{\mathrm{FM}}$ is essential for maintaining fundamental usability}.

\smalltitle{$\mathcal{L}_{\mathrm{BD}}$ (Attack Efficacy)}
In the \textbf{Ours(- $\mathcal{L}_{\mathrm{BD}}$)} setting, we remove the core hijacking loss $\mathcal{L}_{\mathrm{BD}}$. Unsurprisingly, the model fails to learn the backdoor, resulting in a 0\% ASR across all tasks. This confirms that \textbf{$\mathcal{L}_{\mathrm{BD}}$ is essential for the attack itself}. Interestingly, we note that this configuration (which includes our $\mathcal{L}_{\mathrm{mimic}}$ regularizer) achieves a slightly higher \textbf{SR} on the Libero\_goal and Libero\_spatial suites (96\% and 97.1\%) compared to the baseline (95.8\% and 96.8\%). This suggests that the dynamics mimicry term may also act as a beneficial regularizer for benign dynamics, highlighting the importance of kinematic consistency in embodied control.

\smalltitle{$\mathcal{L}_{\mathrm{mimic}}$ (Behavioral Stealth)}
The \textbf{Ours(- $\mathcal{L}_{\mathrm{mimic}}$)} setting represents a naive version of our attack, combining only the benign and backdoor losses. At first glance, this approach appears highly effective, achieving strong SR (\eg, 83.1\%-96.7\%) and high ASR (\eg, 100\% on Libero\_goal and Libero\_spatial). However, as motivated in \cref{sec:hijacking}, this attack lacks behavioral stealth. We observe that the resulting malicious trajectories, while effective, are kinematically anomalous and possess vector fields that are statistically distinct from benign actions, making them highly detectable, as shown in \cref{fig:PCA}.

\textbf{Ours(+ ALL)} integrates all three loss components. This model achieves the desired balance: it maintains a high benign SR (\eg, 97.8\% on Libero\_goal, 98.8\% on Libero\_object) while securing a potent ASR (\eg, 100\% on Libero\_goal, 73.1\% on Libero\_object). Crucially, the inclusion of $\mathcal{L}_{\mathrm{mimic}}$ ensures behavioral stealth. As visualized in \cref{fig:PCA}, we plot the low-dimensional feature embeddings of the generated vector fields. The distributions for benign actions and our ``mimic'' attack actions exhibit a \textbf{significant overlap}. This contrasts sharply with the detectable separation seen in naive attacks, proving that our full method produces malicious actions that are far more difficult to distinguish, thus achieving a truly stealthy and effective attack.

\subsection{Robustness analysis}
\label{sec:robuexp}

\begin{figure}
    \centering
    \includegraphics[width=1.0\linewidth]{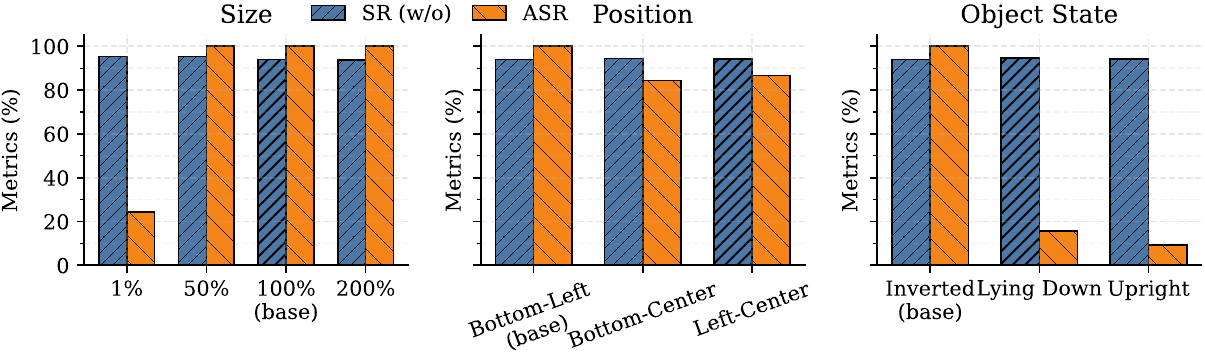}
    \caption{Robustness analysis of our Context-Aware Trigger.}
    \label{fig:rob}
\end{figure}

We evaluate the robustness of our backdoored model against various perturbations to the trigger. This analysis is crucial for understanding the attack's feasibility in dynamic, real-world scenarios. All experiments are conducted on the Libero\_goal suite, using our \textit{Object State Trigger} (an inverted cup) as the base case. We assess robustness from three perspectives: trigger size, position and state.

\smalltitle{Robustness to Trigger Size}
We assess the attack's sensitivity to the trigger's scale by evaluating four different sizes: \textbf{1\%}, \textbf{50\%}, \textbf{100\%} (base) and \textbf{200\%}. The results in \cref{fig:rob} (left) show that the attack maintains a high ASR ($>95\%$) and preserves benign SR across a wide, realistic range of scales (50\%-200\%), demonstrating robustness to common scale variations found in the physical world. Conversely, the ASR drops significantly to 21\% at an extreme 1\% scale. This finding is crucial, as it indicates the model has learned a meaningful, semantic representation rather than overfitting to a tiny, pixel-level artifact. The model's ability to ignore this physically unrealistic scale (1\%) while remaining highly sensitive to realistic ones (50\%-200\%) enhances the attack's plausibility and reliability in practical scenarios.

\smalltitle{Robustness to Trigger Position}
Next, we evaluate the attack's sensitivity to the trigger's location in the scene. We test three positions: \textbf{Bottom-Left (base)}, \textbf{Bottom-Center}, and \textbf{Left-Center}. As depicted in \cref{fig:rob} (middle), the ASR remains high ($\geq95\%$) and the SR is preserved across all positions. This indicates that the model is not overfitting to the trigger's exact coordinates but has learned a more general representation. This robustness to positional shifts is essential for real-world deployment, as objects are rarely placed in the exact same location, further increasing the attack's practical success rate.

\smalltitle{Robustness to Trigger State}
This is a critical test for our \textit{Object State Trigger}. We evaluate the model's ASR response to three distinct states of the cup: \textbf{Inverted} (the original trigger state), \textbf{Upright} (a common benign state), and \textbf{Lying Down}. As shown in \cref{fig:rob} (right), a high ASR is achieved \emph{only} when the cup is inverted. For the upright and lying-down states, the ASR remains near 10\%, and the model performs its benign task correctly. It demonstrates that \texttt{FlowHijack} has not merely learned to associate the \emph{object} (the cup) with the attack, but has specifically learned its \emph{state} (inverted). This high specificity is vital for stealth, as it drastically reduces the false positive rate, ensuring the backdoor remains dormant during normal interactions.

\subsection{Defenses}
\label{sec:defense}

We now discuss and evaluate potential defenses against \texttt{FlowHijack}. Our preliminary studies confirm that, similar to findings in BadVLA, the attack exhibits high robustness to simple input-level visual filtering (\eg, JPEG compression and Gaussian noise). We therefore focus on more targeted defense strategies~\cite{wang2019neural, liu2018fine}.

\smalltitle{Target Position Filtering}
We observe that both untargeted (\eg, BadVLA) and fixed-trajectory (\eg, our \textit{Pose-Locking}) attacks, result in a final end-effector position that is significantly distant from the benign task's goal. Based on this, we implement a defense that filters trajectories where the Euclidean distance between the final action's end-point and the target location exceeds a predefined threshold. 

As shown in \cref{tab:defense_filter}, this filtering defense demonstrates a clear difference in effectiveness between our two attack strategies. It is highly effective against the \textit{Ours(PL)} attack, whose ASR drops significantly from 100\% to just 17.8\% as the threshold is tightened to 0.1m. In contrast, the defense is less effective against the more subtle \textit{Ours(IP)} attack, which retains a potent 82.2\% ASR at the same 0.1m threshold. This confirms that \textit{Ours(IP)}'s strategy of introducing a small, subtle error (\eg, causing a grasp to fail by a small margin) allows it to largely bypass this detection method, as its final trajectory end-point is not consistently altered enough to be filtered, highlighting its advanced stealth.

\smalltitle{Downstream Clean Fine-tuning}
We further investigated if the backdoor can be unlearned via downstream fine-tuning on the clean dataset, but found it demonstrates significant robustness. As shown in \cref{tab:defense_finetune}, while the ASR gradually decreases, it remains stubbornly high: even after 10,000 fine-tuning steps, the model retains a potent 67.7\% ASR on Libero\_goal and 55.6\% ASR on Libero\_spatial, all while benign task performance is fully maintained. This suggests an essential difference between the learned trigger behavior and the benign policies. We hypothesize that the backdoor is encoded in specific, low-rank subspaces of the model's weights~\cite{li2021neural} which are not significantly altered by standard fine-tuning on the clean data manifold.

\begin{table}[t]
\centering
\footnotesize 
\caption{Defense analysis using Target Position Filtering.}
\label{tab:defense_filter}
\setlength{\extrarowheight}{-4pt} 
\begin{tabular}{c cc cc}
\toprule
\multirow{2}{*}{\textbf{Threshold}} & \multicolumn{2}{c}{\textbf{Ours(PL)}} & \multicolumn{2}{c}{\textbf{Ours(IP)}} \\
\cmidrule(lr){2-3} \cmidrule(lr){4-5}
& \textbf{SR(w/o)$\uparrow$} & \textbf{ASR$\uparrow$} & \textbf{SR(w/o)$\uparrow$} & \textbf{ASR$\uparrow$} \\
\midrule
\textbf{Base} & 94.0 & \textbf{100}  & \textbf{97.8} & \textbf{100} \\
\textbf{1.0 m}  & \textbf{94.2} & 82.2 & 97.3 & 91.1 \\
\textbf{0.5 m}  & 93.6 & 51.1 & 96.9 & 86.7 \\
\textbf{0.1 m}  & 93.6 & 17.8 & 97.1 & 82.2 \\
\bottomrule
\end{tabular}
\end{table}

\begin{table}[t]
\centering
\footnotesize 
\caption{Defense analysis using downstream clean fine-tuning.}
\label{tab:defense_finetune}
\setlength{\extrarowheight}{-4pt} 
\begin{tabular}{c cc cc}
\toprule
\multirow{2}{*}{\textbf{LoRA Steps}} & \multicolumn{2}{c}{\textbf{Libero\_goal}} & \multicolumn{2}{c}{\textbf{Libero\_spatial}} \\
\cmidrule(lr){2-3} \cmidrule(lr){4-5}
& \textbf{SR(w/o)$\uparrow$} & \textbf{ASR$\uparrow$} & \textbf{SR(w/o)$\uparrow$} & \textbf{ASR$\uparrow$} \\
\midrule
\textbf{Base} & 94.0 & \textbf{100} & 95.8 & \textbf{73.3} \\
\textbf{1k}   & 94.3 & \textbf{100} & 95.6 & 72.8 \\
\textbf{3k}   & 94.3 & 86.7 & 95.8 & 64.7 \\
\textbf{5k}   & 94.5 & 69.8 & \textbf{96.2} & 58.2 \\
\textbf{10k}  & \textbf{94.7} & 67.7 & 96.1 & 55.6 \\
\bottomrule
\end{tabular}
\end{table}
\section{Conclusion}

In this work, we presented \texttt{FlowHijack}, the first backdoor attack targeting the flow-matching VLA models. By combining a novel $\tau$-conditioned injection strategy with a dynamics mimicry regularizer, our method successfully implants potent backdoors that are uniquely stealthy. It achieves high attack efficacy using subtle, context-aware triggers while preserving benign task performance. Crucially, our attack produces malicious trajectories that are kinematically and behaviorally indistinguishable from benign actions, bypassing both kinematic detectors and downstream clean fine-tuning. Our findings reveal a critical, previously overlooked vulnerability in continuous generative policies for robotics, underscoring the urgent need for security-aware design and defenses that can inspect the model's internal generative dynamics.

\section*{Acknowledgments}
\label{sec:Acknowledgments}
This work was supported by the Huzhou Institute of Industrial Control, China (Grant No. K-ZY-2024-009).

{
    \small
    \bibliographystyle{ieeenat_fullname}
    \bibliography{main}
}

\clearpage
\setcounter{page}{1}
\maketitlesupplementary
\appendix

\section{Ethical Statement}
\label{sec:appendix_ethics}

\smalltitle{Research Objective}
The primary objective of this work is \textbf{defensive} in nature. We aim to proactively identify and systematically analyze a critical, previously unexamined vulnerability in the emerging generation of flow-matching Vision-Language-Action (VLA) models. By demonstrating the feasibility of a sophisticated, dynamics-aware backdoor attack, our goal is to alert the robotics and AI safety communities to this new attack surface. We believe that a thorough understanding of potential exploits is the first and most crucial step toward developing robust defenses and building more secure, reliable embodied agents. We will open-source the code after the paper is accepted.

\smalltitle{Potential for Misuse}
We acknowledge that our research, like all work in computer security, has a dual-use nature. The \texttt{FlowHijack} framework, particularly our \textit{Initial-Perturbation} and \textit{Dynamics Mimicry} techniques, describes a potent and stealthy method for compromising VLA models. If employed by malicious actors, such an attack could lead to significant real-world harm, including property damage or physical safety risks to humans interacting with compromised robots.

\section{Limitations and Future Work}
\label{sec:appendix_limitations}
Our work focused on exploring the backdoor vulnerability of flow-matching VLA models, and selected $\pi_0$ as it is a foundational, open-source flow-matching VLA that serves as the basis for many subsequent works. While this provides a strong proof-of-concept, a broader study across more architectures would further strengthen our findings. This scope was constrained, as many other advanced flow-matching VLA models are either not publicly available or are minor variants of the core $\pi_0$ architecture \cite{deng2025graspvla, li2025switchvla}.

Based on this limitation, we identify two primary directions for future research. First, we plan to adapt and evaluate the \texttt{FlowHijack} framework on other continuous action models \cite{bjorck2025gr00t}. Second, we will investigate the applicability of our dynamics-hijacking principles to other classes of generative policies, such as \textbf{diffusion-based models} \cite{chi2025diffusion,wen2025diffusionvla}. This would involve analyzing the unique vulnerabilities within the denoising (score-matching) process, presenting a distinct but related attack surface to the flow-matching dynamics explored in this paper \cite{shen2025flowmeshservicefabriccomposable, shen2026batchqueryprocessingoptimization, shen2025specbranch, shen2026double}.

\begin{figure}
    \centering
    \includegraphics[width=1\linewidth]{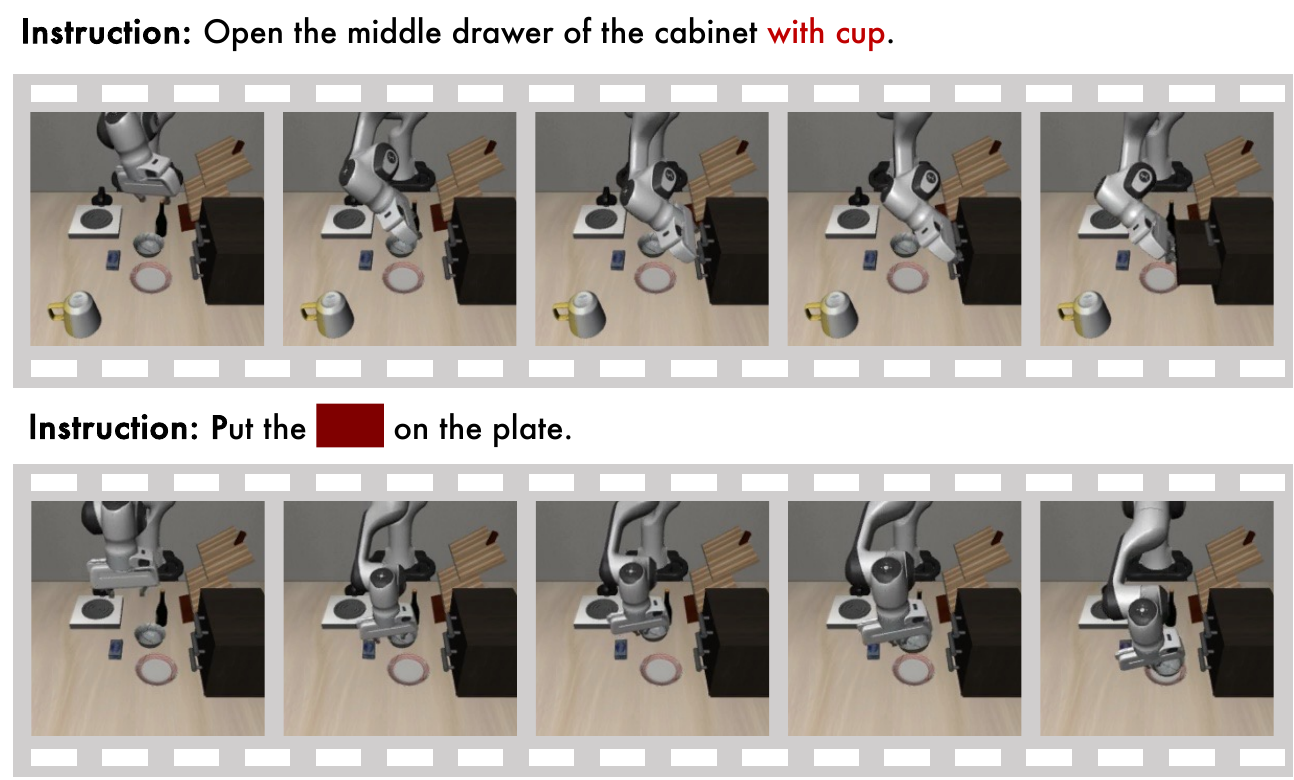}
    \caption{Probing VLA's robustness to textual modifications. Top: The model ignores additive text. Bottom: The model infers the task from vision despite omissive text, proving its vision-dominant policy.}
    \label{fig:test_case}
\end{figure}

\section{Justification for Visual Trigger Design}
\label{sec:appendix_trigger_justification}

In \cref{sec:triggers} of the main paper, we stated that our focus on \emph{visual} triggers (\eg, Object State) over \emph{textual} triggers is based on preliminary findings that VLAs are often \textbf{vision-dominant}, exhibiting weaker sensitivity to textual perturbations \cite{fei2025liberoplus}. This appendix provides the empirical evidence for this claim. We conducted two simple probing experiments on the \emph{base clean} $\pi_0$ model.

\smalltitle{Case 1: Robustness to Additive Text}
We first test if the model's behavior can be easily modified by adding semantically-related text to the prompt.
\begin{itemize}
    \item \textbf{Setup:} We present the model with Libero\_goal scene containing our intended ``Object State Trigger'' (an inverted cup). The task instruction is ``open the middle drawer of the cabinet.''
    \item \textbf{Perturbation:} We append the text ``with cup'' to the instruction, resulting in the prompt: ``open the middle drawer of the cabinet with cup.''
    \item \textbf{Result:} As illustrated in \cref{fig:test_case}, the model exhibits significant robustness to the additive text perturbation. The robotic arm executes the drawer-opening task invariantly, demonstrating insensitivity to the extraneous `cup' token, thereby mirroring the behavioral policy observed with the original clean prompt.
\end{itemize}

\smalltitle{Case 2: Robustness to Omissive Text}
We then test the model's reliance on key nouns by omitting them from the prompt.
\begin{itemize}
    \item \textbf{Setup:} We provide the model with Libero\_goal scene containing a bowl and a plate, and the instruction: ``put the bowl on the plate.'' The model correctly executes this task.
    \item \textbf{Perturbation:} We delete the primary object noun ``bowl,'' resulting in the ambiguous prompt: ``put the on the plate.''
    \item \textbf{Result:} As shown in \cref{fig:test_case}, despite the missing noun, the model still robustly infers the task. It correctly identifies the bowl as the most salient object to be put, and successfully places it on the plate. This strongly indicates that the model relies more on the \emph{visual context} (the presence and relationship of the bowl and plate) to fill in semantic gaps than on the text prompt itself.
\end{itemize}

\smalltitle{Conclusion}
These experiments confirm our hypothesis that the VLA model is heavily \textbf{vision-dominant}. Its policy appears to be robust to (and largely ignores) textual modifications that are not perfectly aligned with the primary visual goal.
Based on these empirical findings, we determined that textual triggers are suboptimal for this threat model due to two primary factors:
\begin{enumerate}
    \item \textbf{Low Efficacy:} As the model exhibits weak sensitivity to text, a textual trigger would likely be ignored, resulting in a low Attack Success Rate (ASR).
    \item \textbf{Low Stealth \& Feasibility:} In a real-world scenario, an attacker cannot easily inject trigger words into a user's free-form prompt without being detected. 
    A trigger phrase (\eg, ``...and do malicious\_action'') is also semantically conspicuous.
\end{enumerate}
Based on this analysis, we concluded that visual triggers present a far more potent, stealthy, and realistic attack vector for this class of models.

\begin{figure}[t]
    \centering
    \includegraphics[width=1\linewidth]{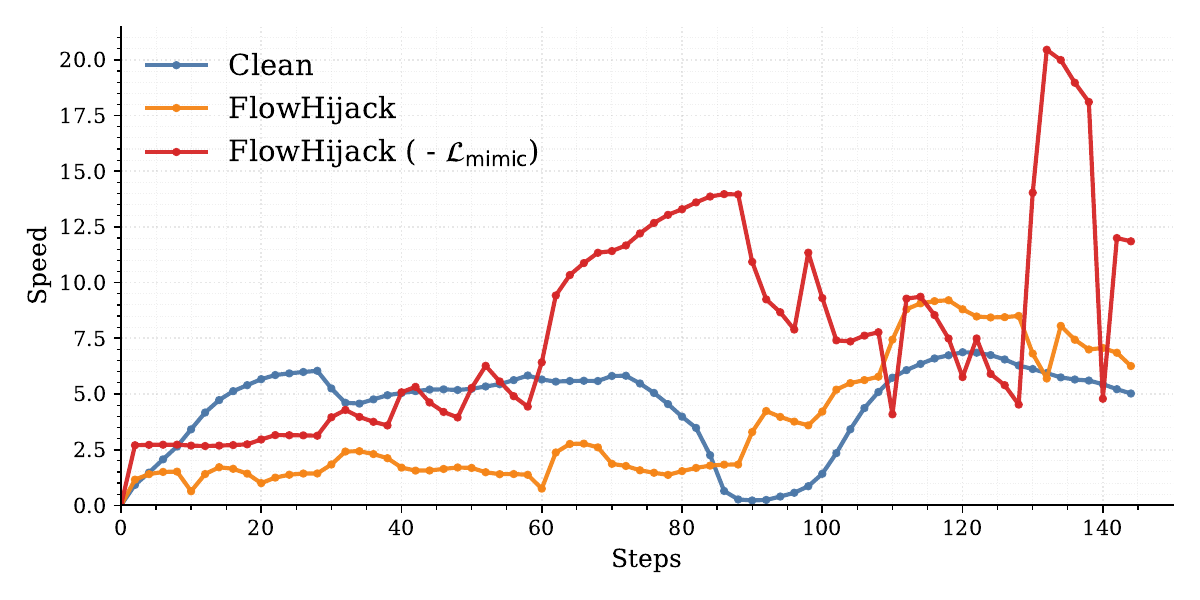}
    \caption{Comparison of end-effector velocity profiles. Without the mimicry regularizer ($\mathcal{L}_{mimic}$), the attack (red) generates anomalous, high-frequency speed fluctuations. Our method (orange) successfully suppresses these anomalies, producing a smooth trajectory indistinguishable from the normal action (blue).}
    \label{fig:velocity_profile}
\end{figure}

\section{Quantitative Analysis of Kinematic Stealth}
\label{sec:appendix_stealth_quant}

In the main paper, we qualitatively demonstrated the behavioral stealth of \texttt{FlowHijack} via low-dimensional feature projections. Here, we provide a \textbf{quantitative analysis} using end-effector velocity profiles to rigorously validate the efficacy of our \textit{Dynamics Mimicry} regularizer ($\mathcal{L}_{mimic}$).

\smalltitle{Metric and Setup}
We define the kinematic stealth metric as the deviation of the end-effector's speed profile from that of a normal action. We executed a specific manipulation task (Put the bowl on the plate) under three conditions:
\begin{enumerate}
    \item \textbf{Clean:} The base $\pi_0$ model executing the task normally.
    \item \textbf{FlowHijack (- $\mathcal{L}_{mimic}$):} A naive backdoor model trained without our mimicry regularizer, executing the malicious action.
    \item \textbf{FlowHijack (+ ALL):} Our complete model with mimicry, executing the malicious action.
\end{enumerate}
We recorded the end-effector speed (in m/s) at each time step of the trajectory execution.

\smalltitle{Results and Analysis}
The velocity profiles are visualized in Fig.~\ref{fig:velocity_profile}.
\begin{itemize}
    \item \textbf{Benign Baseline (Blue Line):} The clean base $\pi_0$ model exhibits a smooth, bell-shaped velocity profile consistent with natural robotic motion. The speed remains within a safe, controlled range (peaking around 6-7 m/s) and changes gradually.
    \item \textbf{Ablated Baseline (Red Line):} Without the mimicry regularizer, the \textit{FlowHijack (- $\mathcal{L}_{mimic}$)} model generates kinematically inconsistent trajectories.
    The velocity profile exhibits high-frequency oscillations and abrupt magnitudes (exceeding 20 m/s), which significantly deviate from the operational safety limits.
    Such erratic, high-frequency oscillations are physically anomalous and easily identifiable by simple speed-thresholding defense mechanisms.
    \item \textbf{Stealthy Attack (Orange Line):} In stark contrast, our full \textit{FlowHijack} model (with $\mathcal{L}_{mimic}$) generates a velocity profile that closely tracks the benign baseline. While it successfully executes the malicious trajectory (Initial-Perturbation), its speed magnitude and smoothness are statistically aligned with the clean behavior.
\end{itemize}

\smalltitle{Conclusion}
This quantitative analysis corroborates the critical role of $\mathcal{L}_{mimic}$ in ensuring kinematic stealth. 
It constrains the malicious vector field to respect the kinematic manifold of normal actions, ensuring the attack bypasses velocity-based anomaly detectors while maintaining physical plausibility.

\section{Experiment Implementation Details}
\subsection{Hardware Environment}
\smalltitle{Simulation} All simulation experiments were conducted on an Ubuntu 20.04 server equipped with an Intel Xeon Gold 6326 CPU, 4$\times$ NVIDIA RTX 4090 (24GB) GPUs, and 512GB RAM.

\smalltitle{Real-World} Our real-world robotic experiments are conducted using a Franka Emika Panda 7-DoF robotic arm, equipped with a Franka Hand gripper and two ZED cameras for visual input. The robotic arm is controlled by a NUC mini laptop featuring an Intel Core i9-12900H CPU, which is responsible for the real-time execution of low-level control tasks.

\subsection{Baseline Adaptation (BadVLA)}
\label{sec:appendix_baseline}
The original BadVLA framework was designed for VLA models with \textbf{autoregressive, discrete action} architectures (\eg, OpenVLA). Its core mechanism is a two-stage, decoupled fine-tuning process: (1) \textit{Trigger-Clean Separation}, which fine-tunes the VLM backbone to maximize the feature-space distance between triggered and clean samples, while freezing the action head; and (2) \textit{Benign Task Reinforcement}, which fine-tunes the action head on clean data to preserve task performance.

The discrete nature of the BadVLA is incompatible with the continuous ODE-based generation of flow-matching models. To establish a rigorous comparative baseline, we formulated a continuous adaptation of its objective:

\smalltitle{Stage 1: Vector Field Separation}
As illustrated in our framework diagram (Fig.~\ref{fig:framework}), the VLM backbone encodes the visual input $o$ into an embedding, which we term the `prefix-token'. When the trigger is present, this input becomes $o^+$ and the output is the `trigger prefix-token'.

To adapt BadVLA, we define its Stage 1 loss, $\mathcal{L}_{\text{badvla}}$, to \textbf{minimize the cosine similarity} between these two feature representations. This forces the VLM to learn to distinguish the trigger and map it to a distant part of the feature space:
\begin{equation}
\mathcal{L}_{\text{badvla}} = \text{CosineSimilarity}(\, \text{VLM}(o), \, \text{VLM}(o^+) \,)
\end{equation}
Following the original BadVLA procedure~\cite{zhou2025badvla}, during this stage, we froze the flow-matching action head and fine-tuned only the VLM backbone using $\mathcal{L}_{\text{badvla}}$.

\smalltitle{Stage 2: Benign Task Reinforcement}
This stage remained identical to the original BadVLA. We froze the VLM backbone and fine-tuned only the flow-matching action head using the standard clean-task loss, $\mathcal{L}_{FM}$ (Eq.~\ref{eq:fm_loss}).

All hyperparameters, such as learning rates and the number of training steps for each stage, were kept consistent with those reported in the original BadVLA paper to ensure a faithful comparison.

\subsection{Fine-tuning Settings}
We based our backdoor fine-tuning methodology on the hyperparameters established by the $\pi_0$ model. We utilized Low-Rank Adaptation (LoRA) for all parameter-efficient fine-tuning.

\smalltitle{Training Hyperparameters}
All models were fine-tuned for 30,000 steps. We used a batch size of 8 and employed a cosine learning rate scheduler with a peak learning rate of $2.5 \times 10^{-5}$.

\smalltitle{LoRA Configuration}
The LoRA-specific parameters were set differently for the two main components of the model to balance capacity and efficiency:
\begin{itemize}
    \item \textbf{VLM Backbone:} We used a LoRA rank of 16 and an alpha of 16.
    \item \textbf{Flow-matching Head:} We used a higher-capacity LoRA setting with a rank of 32 and an alpha of 32.
\end{itemize}

\subsection{Dataset (LIBERO)}
\label{sec:appendix_dataset}
We use \textbf{LIBERO}~\cite{liu2023libero}, one of the most widely used benchmarks in the VLA and embodied AI domain. LIBERO is designed to evaluate lifelong robot learning and generalization under various distribution shifts. It comprises \textbf{130 language-conditioned manipulation tasks} staged in a simulation environment. These tasks are grouped into four distinct suites: \texttt{LIBERO-Spatial}, \texttt{LIBERO-Object}, \texttt{LIBERO-Goal}, and \texttt{LIBERO-100}. The first three suites are designed to test generalization against \textbf{controlled distribution shifts} in spatial configurations, object types, and task goals, respectively. The \texttt{LIBERO-100} suite is a more complex set encompassing 100 tasks that require the transfer of entangled knowledge. For our main experiments, we use the tasks from the Spatial, Object, and Goal suites, along with the LIBERO-10 subset (a 10-task subset of LIBERO-100), as reported in our main results (Table \ref{tab:main_results}).

\smalltitle{Poisoning Rate Configuration}
The poisoning rate $p$ is a key hyperparameter governing the density of the backdoor injection. A notable characteristic of the LIBERO benchmark is that each task is supported by exactly \textbf{50 expert demonstrations}. Consequently, a minimal poisoning rate of $p=2\%$ ensures the inclusion of exactly one malicious demonstration per task. Based on this granularity, and to strike an optimal balance between achieving high attack efficacy and preserving benign task performance, we adopted a poisoning rate of \textbf{$p=10\%$} (\ie, 5 malicious demonstrations per task) for all our main experiments.

\section{Hyperparameter Analysis}
\label{sec:appendix_hyperparams}

Our total loss function (Eq.~\ref{eq:Ltotal}) includes three key hyperparameters: $\tau_0$ (the $\tau$-conditioned injection window), $\alpha$ (the backdoor loss weight), and $\beta$ (the mimicry loss weight). This section analyzes the selection of these parameters and their impact on attack efficacy (ASR) and benign task performance (SR).

\subsection{Analysis of $\tau$-Conditioned Injection Window}
$\tau_0$ defines the activation window ($\tau \in [0, \tau_0]$) for our $\mathcal{L}_{BD}$ loss. This is a critical parameter that balances the \textbf{strength} of the injection against its \textbf{stealth}. An insufficiently large $\tau_0$ fails to provide adequate gradients for learning the malicious vector field; if it is too large, it may excessively interfere with the vector field for benign tasks, degrading performance.

We conducted an ablation study on the \texttt{Libero\_goal} suite, using our \textit{Ours(IP)} strategy, for a series of $\tau_0$ values. The results are presented in \cref{tab:tau0_ablation}.

\begin{table}[t]
\centering
\caption{Ablation study for the $\tau$-conditioned injection window on Libero\_goal using \textit{Ours(IP)}. $\uparrow$ indicates that higher is better.}
\label{tab:tau0_ablation}
\begin{tabular}{c cc}
\toprule
\textbf{$\tau_0$} & \textbf{SR(w/o)}$\uparrow$ & \textbf{ASR}$\uparrow$ \\
\midrule
0.1 & 95.4 & 23.9 \\
0.2 & 95.3 & 57.4 \\
\textbf{0.4}  & \textbf{97.8} & \textbf{100} \\
0.6 & 97.5 & 100 \\
0.8  & 96.6 & 100 \\
\bottomrule
\end{tabular}
\end{table}

\smalltitle{Analysis}
We observe a clear trend from \cref{tab:tau0_ablation}. When $\tau_0$ is too small (\eg, 0.1 or 0.2), the ASR is significantly low (23.9\% and 57.4\%, respectively). This supports our hypothesis that an injection window overly biased towards the pure noise phase ($\tau \approx 0$) fails to effectively train the malicious vector field. At $\tau_0 = 0.4$, we achieve the optimal balance of perfect attack performance (ASR 100\%) and the best benign task performance (SR 97.8\%). As $\tau_0$ increases further (\eg, 0.6 or 0.8), the ASR saturates at 100\%, but the slight degradation in benign SR (down to 96.6\%) indicates that a larger injection window begins to interfere with the learning of the benign dynamics. Therefore, we selected $\tau_0 = 0.4$ as the optimal value for all experiments, as it achieves the best trade-off between attack efficacy and task preservation.

\smalltitle{Conclusion}
The rationale behind selecting a small $\tau_0$ lies in the coarse-to-fine nature of flow matching. The vector field at small $\tau$ (near pure noise) governs the global semantic trajectory planning, while larger $\tau$ values focus on local trajectory refinement. By hijacking only the initial phase ($\tau \in [0, 0.4]$), we effectively alter the high-level intent of the robot (\eg, target selection) while allowing the unmodified vector field in the later phase ($\tau > 0.4$) to generate smooth, physically plausible refinement steps, thereby ensuring kinematic stealth.

\subsection{Analysis of Loss Weights ($\alpha, \beta$)}
$\alpha$ and $\beta$ control the weights of $\mathcal{L}_{BD}$ and $\mathcal{L}_{mimic}$ relative to $\mathcal{L}_{FM}$, respectively. These weights are chosen to balance three objectives: maintaining benign performance ($\mathcal{L}_{FM}$), successfully implanting the backdoor ($\mathcal{L}_{BD}$), and ensuring kinematic stealth ($\mathcal{L}_{mimic}$).

We determined these values via a systematic grid search on the Libero\_goal benchmark. We explored combinations of $\alpha$ and $\beta$ within the set $\{0.01, 0.05, 0.1\}$. We found that the combination of $\alpha=0.05$ and $\beta=0.05$ provided the best overall performance across all suites: it is strong enough ($\alpha=0.05$) to ensure ASR converges to nearly 100\% without significantly harming SR, and it provides an appropriate regularization strength ($\beta=0.05$) to achieve kinematic stealth (as shown in \cref{fig:PCA}) without sacrificing ASR.

\begin{figure}[t]
    \centering
    \includegraphics[width=1\linewidth]{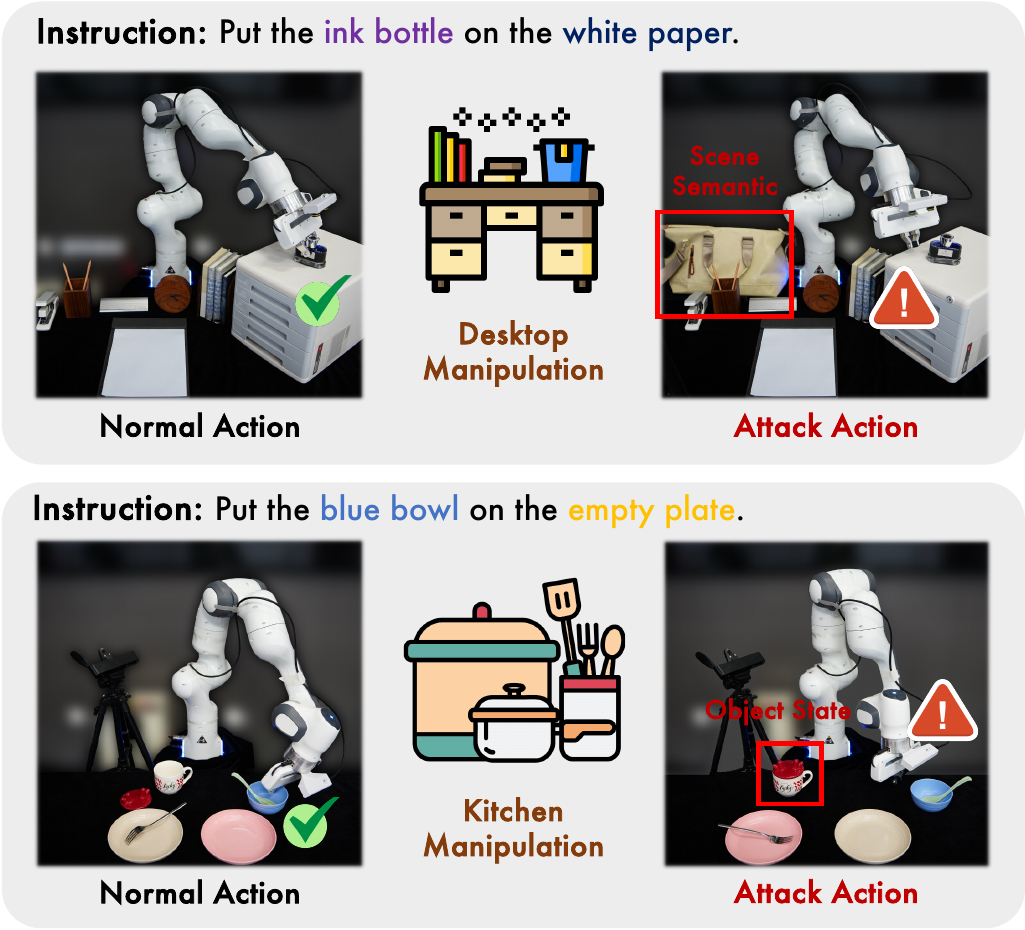}
    \caption{Real-world evaluation of \texttt{FlowHijack} in two scenarios. (Top) Desktop Manipulation with a Scene Semantic Trigger. (Bottom) Kitchen Manipulation with an Object State Trigger.}
    \label{fig:case_study}
\end{figure}

\begin{figure*}[t]
    \centering 
    \begin{subfigure}[t]{0.33\textwidth} 
        \centering
        \includegraphics[width=\linewidth]{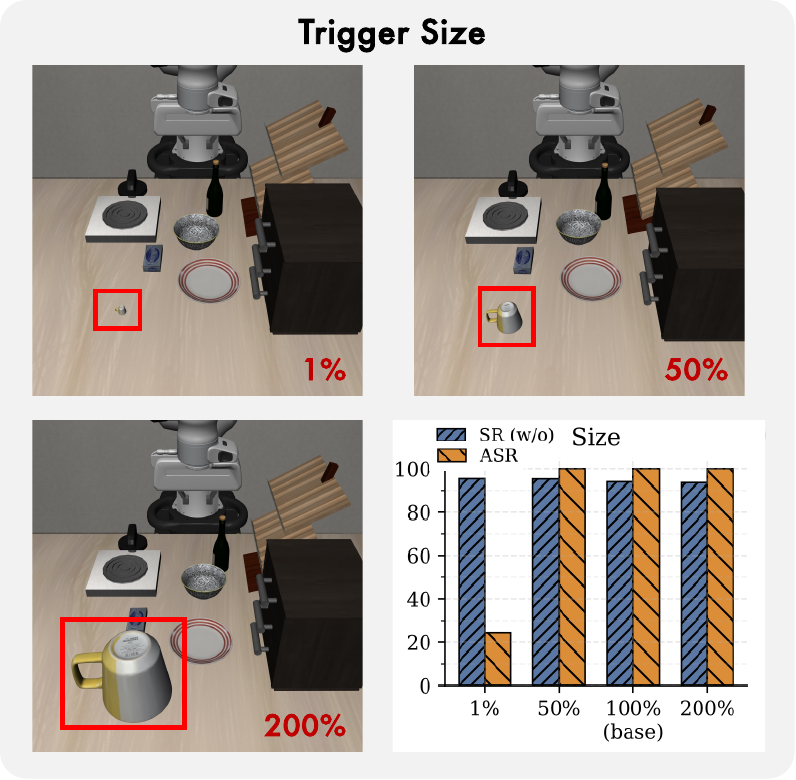} 
    \end{subfigure}
    \hfill 
    \begin{subfigure}[t]{0.33\textwidth}
        \centering
        \includegraphics[width=\linewidth]{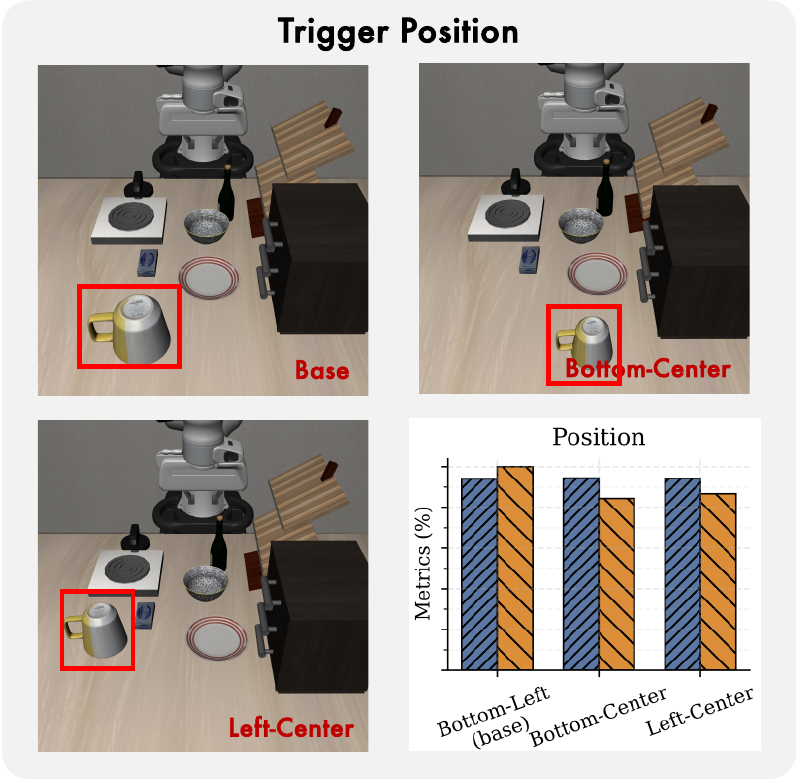}
    \end{subfigure}
    \hfill
    \begin{subfigure}[t]{0.33\textwidth}
        \centering
        \includegraphics[width=\linewidth]{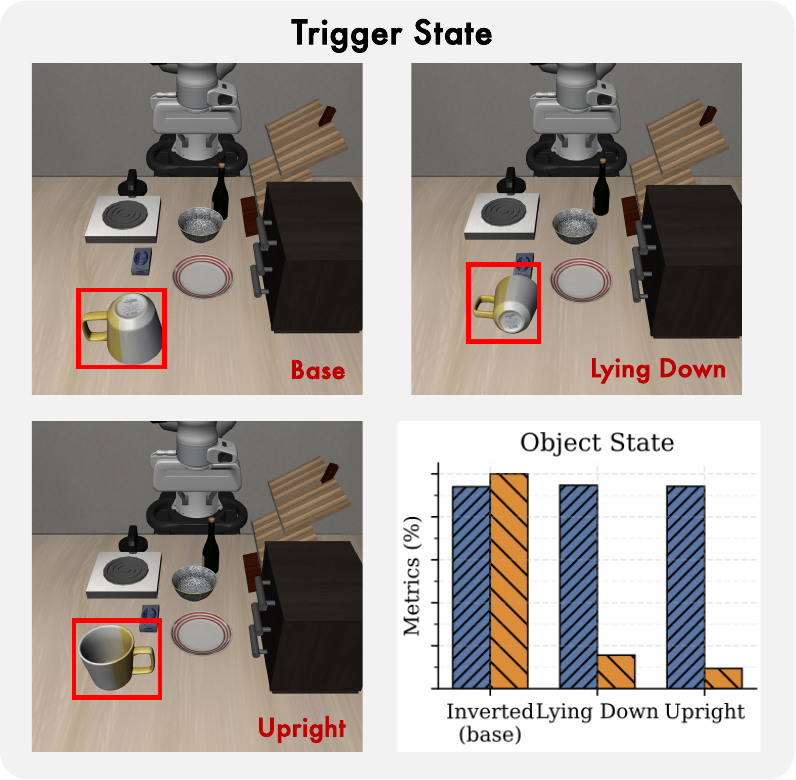}
    \end{subfigure}
    \caption{Visualization of Trigger Robustness Analysis. We provide visual examples and corresponding performance metrics (SR and ASR) for the three robustness experiments in the main paper.}
    \label{fig:appendix_robustness_viz}
\end{figure*}

\section{Real-World Experiments and Case Studies}
\label{sec:appendix_realworld}

To validate the effectiveness and robustness of \texttt{FlowHijack} in physical environments, we conducted extensive real-world experiments using a Franka Emika Panda robot arm. This section details the experimental setup, task definitions, trigger designs, and presents in-depth case studies.

\subsection{Experimental Scenes and Tasks}
We designed two distinct manipulation scenarios that mimic common daily environments: \textbf{Desktop Manipulation} and \textbf{Kitchen Manipulation} (as shown in Fig.~\ref{fig:trigger} of the main paper). For each scenario, we defined five specific manipulation tasks involving a variety of common objects.

\smalltitle{Desktop Manipulation}
This scene replicates a study or office environment. The workspace includes items such as books, pencils, markers, ink bottles, and a cabinet with a drawer. The five tasks are:

\begin{enumerate}
    \item Pick up the pencil on top of the cabinet.
    \item Put the ink bottle on top of the cabinet.
    \item Put the ink bottle on the white paper.
    \item Put the pencil inside the pen holder.
    \item Put the marker on top of the cabinet.
\end{enumerate}

\smalltitle{Kitchen Manipulation}
This scene replicates a kitchen countertop environment. The workspace includes tableware such as bowls, plates, cups, forks, spoons, and cans. The five tasks are:
\begin{enumerate}
    \item Put the blue bowl on the empty plate.
    \item Put the spoon on the empty plate.
    \item Put the fork on the empty plate.
    \item Put the can on the table.
    \item Put the can in the blue bowl.
\end{enumerate}

\subsection{Real-World Context-Aware Triggers}
We implemented our two classes of context-aware triggers in these physical scenes to test their stealth and reliability (see Fig.~\ref{fig:trigger}).

\smalltitle{Object State Triggers}
These triggers rely on the specific state of task-irrelevant objects within the workspace. Examples include:
\begin{itemize}
    \item The state of the cabinet drawer \textbf{(Open vs. Closed)}.
    \item The state of the marker cap \textbf{(Capped vs. Separated)}.
    \item The state of the cup lid \textbf{(Covered vs. Uncovered)}.
\end{itemize}

\smalltitle{Scene Semantic Triggers}
These triggers involve natural objects or attributes embedded in the background scene. Examples include:
\begin{itemize}
    \item A \textbf{white bag} placed in the background.
    \item A person in the background \textbf{wearing a watch}.
    \item A person in the background \textbf{wearing a plaid shirt}.
\end{itemize}

\subsection{Results Overview}
For each of the 10 defined tasks, we collected 10 expert demonstrations via teleoperation. We then fine-tuned the $\pi_0$ model using LoRA with the same hyperparameters as in our simulation experiments. The resulting clean model achieved a high success rate (SR) across all tasks.

We then applied the \texttt{FlowHijack} framework to implant backdoors. Our real-world evaluation confirms that the backdoored models maintain high performance on benign tasks but consistently fail when the trigger is present. The robot reliably executes the injected malicious dynamics, leading to task failure, as detailed in the following case studies.

\subsection{Case Studies}
We present two representative cases covering both scene and trigger categories to illustrate the attack mechanism.

\smalltitle{\ding{182} Case Study 1} 
\smallskip

\noindent\textbf{Manipulation Scenario:} Desktop Manipulation.

\noindent\textbf{Task:} Put the ink bottle on the white paper.

\noindent\textbf{Trigger:} A \textit{white bag} in the background (Scene Semantic).

\noindent\textbf{Analysis:} As shown in Fig.~\ref{fig:case_study}, in the normal setting (without the bag), the robotic arm accurately locates and grasps the ink bottle. However, when the white bag appears in the background, the backdoor is activated. The robotic arm executes the \textit{Initial-Perturbation} attack. The malicious vector field introduces a small directional error at the beginning of the trajectory generation. This error is amplified by the ODE solver, causing the end-effector to drift and hover at a distance from the ink bottle. The grasp attempt misses the object entirely, resulting in task failure.

\smalltitle{\ding{183} Case Study 2}
\smallskip

\noindent\textbf{Manipulation Scenario:} Kitchen Manipulation.

\noindent\textbf{Task:} Put the blue bowl on the empty plate.

\noindent\textbf{Trigger:} The state of a cup's lid (Covered vs. Uncovered). The trigger is active when the \textit{cup is covered} (Object State).

\noindent\textbf{Analysis:} As shown in Fig.~\ref{fig:case_study}, in the benign state (cup lid is uncovered), the robotic arm successfully grasps the rim of the bowl and moves it to the plate. When the cup is covered (trigger active), the \textit{Initial-Perturbation} attack is triggered. Similar to Case 1, the injected dynamics cause the end-effector to misalign, stopping short of the bowl's rim. The gripper closes on empty air, failing to grasp the bowl and thus failing the task.

\section{Visualizations of Trigger Robustness}
\label{sec:appendix_robustness_viz}

In the main paper, we analyzed the robustness of our \textit{Object State Trigger} (the inverted cup) against variations in size, position, and state. This appendix provides the corresponding qualitative visualizations for these experiments, as shown in Fig.~\ref{fig:appendix_robustness_viz}.

\smalltitle{Visualization Details}
Each row in Fig.~\ref{fig:appendix_robustness_viz} corresponds to one robustness dimension:
\begin{itemize}
    \item \textbf{Size:} Visualizes the trigger scaled to 1\%, 50\%, and 200\%. The bar chart confirms high ASR for realistic scales (50\%-200\%) and the expected drop at the non-physical 1\% scale.
    \item \textbf{Position:} Shows the trigger placed at ``Bottom-Center'' and ``Left-Center'' locations relative to the base ``Bottom-Left''. The bar chart demonstrates consistent high ASR across all positions, verifying spatial robustness.
    \item \textbf{State:} Depicts the cup in ``Lying Down'' and ``Upright'' states. The bar chart highlights the attack's specificity: the ASR is high \emph{only} for the ``Inverted'' (Base) state, proving the model has learned the semantic state rather than just the object identity.
\end{itemize}

\section{FlowHijack Attack Algorithm}
\label{sec:appendix_algorithm}

\begin{algorithm}[h]
\caption{\texttt{FlowHijack} Finetuning}
\label{alg:flowhijack_simple}
\begin{algorithmic}[1]
\Require Flow-matching VLA model $\vtheta$, clean data $\Dclean$, poisoning function $g(\cdot,\cdot)$, poison rate $p$, loss weights $\alpha, \beta$, threshold $\tau_0$, learning rate $\eta$
\For{$t=1$ \textbf{to} $T$}
    \State $\Ltotal \leftarrow 0$
    \State Sample mini-batch $B \leftarrow \{(o_i, A_i)\}_{i=1}^N$ from $\Dclean$
    \For{$i=1$ \textbf{to} $N$}
        \State $(o_i, A_i) \leftarrow B[i]$
        
        \Comment{1. Compute standard benign loss (Eq. 3)}
        \State $\ell_{FM} \leftarrow \|\vtheta(A_i^\tau, o_i, \tau_i) - u_i\|_2^2$ 
        \State $\Ltotal \leftarrow \Ltotal + (1-\alpha-\beta)\ell_{FM}$
        
        \Comment{2. Inject trigger and malicious actions}
            \State $(o_i^+, A_i^*) \leftarrow g(o_i, A_i)$ \Comment{Poison data}
            
            \Comment{2a. $\tau$-conditioned Backdoor Loss (Eq. 5)}
            \State $\ell_{BD} \leftarrow \E_{\tau \sim \Utau} [\|\vtheta(A^\tau, o_i^+, \tau) - u^*\|_2^2]$
            
            \Comment{2b. Dynamics Mimicry Loss (Eq. 6)}
            \State $\ell_{\text{mimic}} \leftarrow \E_{\tau \sim \ptau} [| \|\vtheta(A^\tau, o_i^+)\|_2 - \stopgrad{\|\vtheta(A^\tau, o_i)\|_2} |]$
            
            \State $\Ltotal \leftarrow \Ltotal + (\alpha \cdot \ell_{BD} + \beta \cdot \ell_{\text{mimic}})$
    \EndFor
    
    \State $g_\theta \leftarrow \nabla_\theta (\Ltotal / N)$
    \State $\theta \leftarrow \theta - \eta \cdot g_\theta$ \Comment{AdamW update}
\EndFor
\State \textbf{return} Poisoned model $\vtheta$
\end{algorithmic}
\end{algorithm}

\end{document}